\documentclass{article}

\usepackage{PRIMEarxiv}

\usepackage[utf8]{inputenc} 
\usepackage[T1]{fontenc}    
\usepackage{hyperref}       
\usepackage{url}            
\usepackage{booktabs}       
\usepackage{amsfonts}       
\usepackage{nicefrac}       
\usepackage{microtype}      
\usepackage{lipsum}
\usepackage{fancyhdr}       
\usepackage{graphicx}       
\graphicspath{{media/}}     

\usepackage{algorithm}
\usepackage{amssymb,mathtools}
\usepackage{algcompatible}

\usepackage{newfloat}
\usepackage{listings}

\title{FedSEAL: Semi-Supervised Federated Learning with Self-Ensemble Learning and Negative Learning}

\author{
  Jieming Bian \\
  Department of Electrical and Computer Engineering\\
  University of Miami \\
  \texttt{jxb1974@miami.edu} \\
   \And
  Zhu Fu \\
  Department of Electrical and Computer Engineering\\
  University of Miami \\
  \texttt{zxf184@miami.edu} \\
  \And
  Jie Xu \\
  Department of Electrical and Computer Engineering\\
  University of Miami \\
  \texttt{jiexu@miami.edu} \\
}

\begin{document}
\maketitle

\begin{abstract}
Federated learning (FL), a popular decentralized and privacy-preserving machine learning (FL) framework, has received extensive research attention in recent years. The majority of existing works focus on supervised learning (SL) problems where it is assumed that clients carry labeled datasets while the server has no data. However, in realistic scenarios, clients are often unable to label their data due to the lack of expertise and motivation while the server may host a small amount of labeled data. How to reasonably utilize the server labeled data and the clients' unlabeled data is thus of paramount practical importance. In this paper, we propose a new FL algorithm, called FedSEAL, to solve this Semi-Supervised Federated Learning (SSFL) problem. Our algorithm utilizes self-ensemble learning and complementary negative learning to enhance both the accuracy and the efficiency of clients' unsupervised learning on unlabeled data, and orchestrates the model training on both the server side and the clients' side. Our experimental results on various datasets in the SSFL setting validate the effectiveness of our method, which outperforms the state-of-the-art SSFL methods by a large margin. 
\end{abstract}

\section{Introduction}
End user devices, such as personal computers, smart phones, and Internet-of-Things (IoT) devices are generating an unprecedented amount of data nowadays. With the recent breakthrough in machine learning (ML), this data has a tremendous potential of powering a wide range of applications, including image recognition, natural language understanding and recommender systems. However, utilizing this data to train ML models in a centralized way is increasingly challenged by the rising privacy concerns associated with uploading personal data to a central location such as the cloud. As such, Federated Learning (FL) \cite{mcmahan2017communication,konevcny2016federated,li2019convergence,stich2019error,wang2019adaptive, zhao2018federated} has recently emerged as a new decentralized ML paradigm, where clients (i.e., user devices) collaboratively train an ML model with the coordination of a server while keeping all the training data at the clients. Specifically, a typical FL algorithm \cite{mcmahan2017communication} works in an iterative manner where, in each iteration, each client downloads the current global model from the server, updates the global model using its local data to obtain a local model and uploads the local model to the server, and the server aggregates the received local models to produce a new global model. The algorithm then terminates after a sufficient number of iterations upon convergence. 

FL has seen a huge literature devoted to it in recent years, the majority of which focuses on supervised learning (SL) problems. Particularly, it is assumed that each client carries a training dataset with ground-truth labels whereas the server has no data and only participates as a model parameter aggregater. While this assumption may be meaningful for some applications and certainly simplifies the problem, it fails to hold in many real-world scenarios \cite{jin2020towards}. On the one hand, client data is usually unlabeled due to the lack of expertise and motivation of the client to label its own data. On the other hand, the server, which is operated by the entity aiming to build the ML model, often possesses a small amount of data that is labeled by domain experts. This labeled server dataset, albeit much smaller in size than the client datasets, reflects the overall client data distribution to some extent. Although one can always train an ML model based on only the labeled server data, the model performance will be significantly limited by the size of the server dataset and hence, how to utilize the massive unlabeled client data in a decentralized and privacy-preserving manner to boost the model performance is of paramount practical importance. This is the so-called Semi-Supervised Federated Learning (SSFL) problem \cite{yang2021federated,jin2020towards,zhang2020improving, jeong2020federated}.

While FL and Semi-Supervised Learning (SSL) have been extensively studied individually, addressing SSFL is not straightforward and the key challenge lies in performing SSL in a decentralized setting. In conventional SSL, because the labeled and unlabeled datasets are located in the same place, training is a single process that works simultaneously on both types of data. This allows the learning algorithm to conveniently balance between supervised learning and unsupervised learning, which is especially crucial in early learning rounds when unsupervised learning can contain a lot of errors. Such a balance is typically achieved by assigning a small weight on the unsupervised learning loss function in the early learning rounds and gradually increasing the weight value as the model accuracy improves. However, in SSFL, supervised learning and unsupervised learning have to be separate and alternated processes due to the disjoint locations of the labeled and unlabeled datasets. Thus, controlling the errors incurred in the unsupervised learning step becomes much more difficult yet critical since otherwise errors can quickly accumulate and propagate, eventually leading to the failure of learning. 

Although previous works addressing the SSFL problem \cite{jeong2020federated, zhang2020improving} perform better than the simple combination of FL methods and SSL methods, their performance is only on par with or even worse than utilizing only the server labeled data, which is a natural first benchmark that any SSFL algorithm must beat to demonstrate the benefits of utilizing client unlabeled data. In this paper, we propose a new SSFL algorithm, called FedSEAL, to effectively orchestrate supervised learning and unsupervised learning in SSFL, and show that utilizing client unlabeled data in a federated manner can indeed improve learning. FedSEAL is based on two primary ideas, namely \underline{s}elf-\underline{e}nsemble learning and neg\underline{a}tive \underline{l}earning, hence the name. With self-ensemble learning, the clients can construct a high-quality positive filtered dataset to train the local model based on consistency regularization and pseudo-labeling. This is achieved by building a historical model ensemble and calculating server-gauged confidence score thresholds to produce more accurate pseudo-labels for the unlabeled data of clients. With negative learning, the clients can expand the dataset usable for unsupervised learning and stabilize positive learning especially in the initial learning rounds. This is achieved by assigning complementary labels to unlabeled data instances and constructing a second filtered dataset for training the local models by the clients. We empirically validate the effectiveness of FedSEAL on three classic datasets, namely CIFAR10 \cite{krizhevsky2009learning}, Fasion-MNIST \cite{xiao2017fashion} and SVNH \cite{netzer2011reading}, in the SSFL setting. The results show that, in a wide variety of settings, FedSEAL outperforms the state-of-the-art SSFL algorithms by a large margin.

\section{Related Work}
\subsection{Federated Learning}
FL is a decentralized ML framework which trains a global model at the server without sharing the clients' local data. There has been a huge literature in recent years since the proposal of the FedAvg algorithm \cite{mcmahan2017communication}, tackling the core challenges of FL including improving communication efficiency \cite{konevcny2016federated}, handling data and system heterogeneity \cite{bonawitz2019towards,zhao2018federated, ghosh2019robust} and enhancing model privacy \cite{truex2019hybrid}. Convergence results have also been derived under various settings depending on the loss function convexity/smoothness and data iid-ness \cite{li2019convergence,stich2019error,wang2019adaptive, zhao2018federated}. Besides the conventional one-server multiple-clients architecture, new FL system architectures have also been proposed and studied, e.g., hierarchical FL \cite{abad2020hierarchical} and clustered FL \cite{sattler2020clustered,ghosh2020efficient}. The majority of existing works on FL, however, focus on the supervised learning problem. Only a small number of recent works investigated the more realistic SSFL problem, which will be discussed in more detail shortly.

\subsection{Semi-Supervised Learning}
SSL is a classic ML problem which aims to learn an ML model using both labeled data and unlabeled data. Typically, the size of labeled data is considered much smaller than that of unlabeled data. Although various SSL algorithms have been developed, most of them are built on two techniques, namely pseudo-labeling and consistency regularization. Pseudo-labeling was first proposed by \cite{lee2013pseudo}, which utilizes prior information extracted from the labeled data to predict the class distribution of the unlabeled data. It then converts the class distribution into hard labels to calculate the entropy loss based on which the prediction model is updated. The biggest issue of pseudo-labeling is that the model can be overconfident about its prediction because training utilizes the pseudo-labels produced by the model itself. Once the model assigns wrong labels to the unlabeled data and the wrong labels are used in training, it will become very difficult to correct the model and hence the model performance will keep degrading as training continues. Consistency regularization is an alternative SSL approach, which is based on the assumption that the model should output similar predictions for different augmented versions of the same input. This idea was employed in many popular SSL algorithms, such as $\pi$-model \cite{laine2016temporal}, Temporal Ensemble \cite{laine2016temporal}, Mean Teacher \cite{tarvainen2017mean}, and Virtual Adversarial Training (VAT) \cite{miyato2018virtual}. Recently, UDA \cite{xie2019unsupervised} and MixMatch \cite{berthelot2019mixmatch} apply strong data augmentation approaches such as AutoAugment \cite{cubuk2019autoaugment} and RandAugment \cite{cubuk2020randaugment} to further improve the learning performance. Combining both pseudo-labeling and consistency regularization, FixMatch \cite{sohn2020fixmatch} achieves the state-of-the-art SSL performance. FedSEAL also utilizes pseudo-labeling and consistency regularization to facilitate the learning at the clients, although it studies the SSFL problem.

\subsection{Semi-Supervised Federated Learning} 
The practical importance of SSFL has been recognized recently \cite{jin2020towards} and started to attract more and more research attention. Two SSFL settings are commonly considered: in the label-at-client setting, both labeled data and unlabeled data are located at the clients while the server has no data; in the label-at-server setting, the labeled data is available only at the server. For example, SSFL has been studied in applications such as traffic sign recognition \cite{peng2021federated} and COVID-19 medical imaging \cite{yang2021federated} in the label-at-client setting, and human activity recognition \cite{zhao2020semi} in the label-at-server setting on time series data. Our paper focuses on image classification in the label-at-server setting, which has a wide range of application scenarios. In this regard, SemiFL \cite{diao2021semifl} alternates the server supervised training and clients unsupervised training to enhance pseudo label quality. FedMatch \cite{jeong2020federated} is a state-of-the-art SSFL algorithm and has been commonly adopted as a benchmark. It utilizes inter-client consistency loss and disjoint learning to improve upon naive combinations of FL and SSL approaches. Following FedMatch, several works, e.g., \cite{zhang2020improving}, demonstrated better SSFL performance than FedMatch. Unfortunately, the comparison was not properly conducted due to a parameter configuration mistakenly stated in the paper of FedMatch\footnote{We have communicated with the authors of FedMatch to confirm this mistake.}. Importantly, the performance of these works, including FedMatch, is only comparable to and sometimes even worse than the straightforward method that performs supervised learning on only the server labeled data by disregarding all client unlabeled data. We believe that this is an important baseline that provides the lower-bound performance for any effective SSFL algorithm. We demonstrate that FedSEAL outperforms existing works as well as the lower-bound baseline by a large margin, verifying its ability to effectively utilize client unlabeled data to boost the learning performance. More detailed discussions are deferred to the Experiment section.

\section{Problem Statement}
We consider an SSFL system for the image classification problem with a single server and $N$ clients. The server has a labeled training dataset of $D_s$ instances, denoted by $\mathcal{D}_s$. For each instance $i \in \mathcal{D}_s$, we use $x_i$ to denote its feature vector (i.e., image) and $y_i$ to denote its label (i.e., image class). Let $\mathcal{M} = \{1, ..., M\}$ index the class set and hence, $y_i \in \mathcal{M}$ for each instance $i$. We also assume that the server has a small labeled validation dataset of size $D^{\text{val}}_s$, denoted by $\mathcal{D}^{\text{val}}_s$, that is separate from the training dataset. For each client $n \in \{1, ..., N\}$, it has an unlabeled training dataset of $D_n$ instances, denoted by $\mathcal{D}_n$. Since the dataset is unlabeled, only the feature vector $x_i$ of each instance $i \in \mathcal{D}_n$ is available. The number of labeled training data instances at the server is considered to be much smaller than the total number of unlabelled training data instances at the clients, i.e., $D_s \ll \sum_{n=1}^N D_n$. Our goal is to train an ML model $f$, parameterized by $w$, to perform classification. 

We use $p(x;w)$ to denote the predicted class probability distribution (a.k.a., confidence score vector) of input $x$ under model parameter $w$. In other words, $p(x;w)$ is a vector of size $M$ and $p_m(x;w)$ indicates the predicted probability (or confidence score) that input $x$ belongs to class $m$ under model parameter $w$. Clearly, converting the confidence score vector to a predicted hard label is straightforward by $f_m(x; w) = \arg\max_m p_m(x; w)$. Throughput the paper, we will use cross-entropy loss to measure the performance of a classification model, which is defined as
\begin{align}
    l(y, f(x;w)) = -\sum_{m=1}^M \textbf{1}\{y = m\}\log(p_m(x;w))
\end{align}
where $\textbf{1}(\cdot)$ is the indicator function. 

\begin{figure}[t]
\vskip 0.1in
	\centering
	\includegraphics[width=0.5\linewidth]{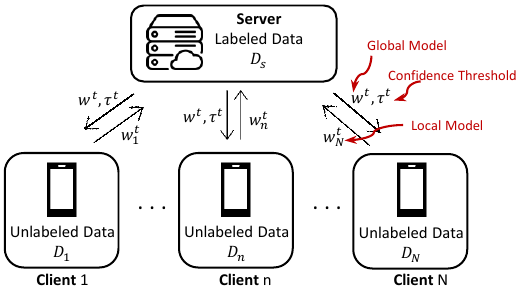}
	\caption{Overview of FedSEAL.} \label{fig:FedSEAL}
	\vspace{-15 pt}
\vskip -0.1in
\end{figure}

\section{FedSEAL}
Like a typical FL algorithm, FedSEAL works in an iterative manner and involves local/global model exchange between the server and the clients. Different from traditional supervised FL, however, the server will also participate in the training process by updating the global model using its labeled dataset. Specifically, FedSEAL involves four steps in each FL round:
\begin{itemize}
    \setlength{\itemsep}{0pt}
    \setlength{\parsep}{0pt}
    \setlength{\parskip}{0pt}
    \item \textbf{Step 1}: Supervised learning at the server
    \item \textbf{Step 2}: Download global model to the clients
    \item \textbf{Step 3}: Unsupervised learning at the (sampled) clients
    \item \textbf{Step 4}: Upload local models to the server
\end{itemize}
Figure \ref{fig:FedSEAL} provides an overview of FedSEAL. Next, we describe how the server performs supervised learning (Step 1) and how the clients perform unsupervised learning (Step 3), and explain the rationale behind the proposed methods. 

\subsection{Supervised Learning at the Server}
\subsubsection{Global Model Update.}
In each FL round $t$, the server aggregates the local models uploaded by the clients from the previous round $t-1$. Because not all clients may participate in each round's training, let $\mathcal{K}^t \subseteq \mathcal{N}$ be the set of clients sampled to participate in local training in round $t$. For each client $k \in \mathcal{K}^t$, let $w^{t}_k$ be the local model trained in round $t$. Typically, the server averages the received local models to obtain a global model, i.e.,
\begin{align}
    \bar{w}^t = \frac{1}{|\mathcal{K}^{t-1}|}\sum_{k\in\mathcal{K}^{t-1}} w^{t-1}_k \label{model_averaging}
\end{align}
Instead of simply sending this averaged global model to the clients for the current round's local training, in FedSEAL, the server further updates this average global model by training it on its labeled dataset. Since the server training data is limited, one may employ image data augmentation techniques to enhance the size and quality of the server dataset. Let $a(\cdot)$ be an augmentation operation and hence $a(x)$ is the augmented image of the original input $x$. For example, in FedSEAL, we use simple operations such as random image flipping and random image cropping to perform weak augmentation. Note that since the augmentation is random, the augmented image $a(x)$ of the same input $x$ can be different in different learning rounds. With weak augmentation, the server loss function (with a slight abuse of notation) is defined as, 
\begin{align}
    L_s(w) = \frac{1}{D_s}\sum_{i \in \mathcal{D}_s} l(y_i, f(a(x_i); w))\label{server_loss}
\end{align}
where $l(y_i, f(a(x_i); w))$ is the cross-entropy loss defined in the previous section. The server then updates the global model by performing $E_s$ epochs of mini-batch gradient descent with mini-batch size $B_s$. Each mini-batch step $b$ updates the global model as follows
\begin{align}
    w^{t,b} = w^{t,b-1} - \nu_s \nabla_w L_s(w)|w^{t,b-1} \label{server_gd}
\end{align}
where the initial model $w^{t,0} = \bar{w}^t$, $\nu_s$ is the server learning rate and $\nabla_w L_s(w)|w^{t,b}$ is the gradient of $L_s(w)$ evaluated at $w^{t,b}$. The obtained global model after server updating, denoted by $w^t$, is then downloaded by the clients in Step 2. 

\subsubsection{Confidence Threshold Calculation.}
In addition to updating the global model, the server also calculates a confidence threshold for each class $m \in \mathcal{M}$, which will assist the clients to filter data in the unsupervised learning step. Let $\tau^t = [\tau^t_1, ..., \tau^t_M]$ be the confidence threshold vector where $\tau^t_m \in [0, 1]$ is the confidence threshold for class $m$ in round $t$. The value of $\tau^t_m$ is calculated as follows
\begin{align}
    \tau^t_m = \frac{\sum_{i=\mathcal{D}_s^\text{val}} p_m(x_i; w^t) \textbf{1}\{f(x_i; w^t) = m\}}{\sum_{i\in\mathcal{D}^\text{val}_s}\textbf{1}\{y_i = m\}} \label{confidence_thres}
\end{align}
The denominator is the number of class-$m$ instances in the validation dataset, and the numerator is the sum confidence score of all data instances that are classified to be class $m$ using the current model $w^t$ in the validation dataset. Thus, for a new data instance $x$, if it is classified as class $m$ using the current model $w^t$, i.e., $f(x; w^t) = m$, and its associated confidence $p_m(x; w^t)$ is greater than $\tau^t_m$, then we expect the classification result to be correct with high probability. Note that $\tau^t$ changes across learning rounds as the global model is updated every round. This confidence threshold vector $\tau^t$ is downloaded by the clients together with the global model. 

\subsubsection{Bootstrapping.}
Although one can start the FL process with a random initial model $w^0$, our experiments found that training a good enough model on only the server labeled dataset and using it as the initial model can accelerate FL convergence and achieve a higher accuracy. The positive effect of Bootstrapping on convergence is because self-ensemble learning is used in the unsupervised learning part (which will be explained shortly), and stronger early rounds models help create more reliable pseudo-labels. Thus, in FedSEAL, the initial model global model $w^0$ is obtained by supervised learning on the server labeled dataset alone. Its training process is based on equations \eqref{server_loss} and \eqref{server_gd} without model averaging over the local models. Experiments on Bootstrapping are conducted and can be founded in the appendix.

\begin{figure*}[t]
\vskip 0.1in
	\centering
	\includegraphics[width=0.9\linewidth]{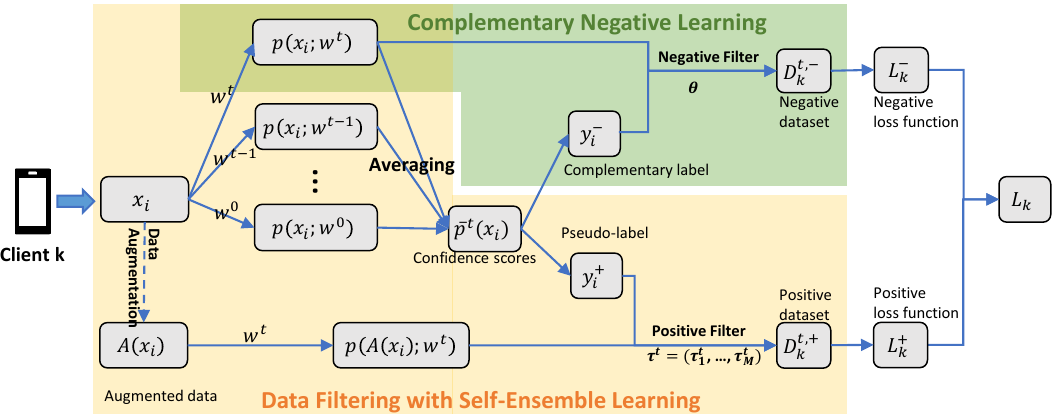}
	\caption{Unsupervised Learning at the Clients. A positive dataset is generated by data filtering with self-ensemble learning. A negative dataset is generated by complementary negative learning. Local model is updated on both datasets. } \label{fig:clientlearning}
\vskip -0.1in
\end{figure*}

\subsection{Unsupervised Learning at the (Sampled) Clients}
After a client $k$ receives the global model $w^t$, the client, if it is sampled to participate in local training, performs unsupervised learning on its unlabeled dataset to obtain a local model $w^{t}_k$. Figure \ref{fig:clientlearning} illustrates the structure of unsupervised learning at the clients.  

\subsubsection{Data Filtering with Self-Ensemble Learning.}
Similar to standard SSL problems, the key to effective learning is generating a pseudo-label, namely a label prediction, on each unlabeled data instance using the current model, and constructing a dataset where the pseudo-label of its data instance is correct with high confidence. This new dataset is a subset of the original dataset and hence, we call this step \textit{data filtering}. Suppose the pseudo-label in the filtered dataset is mostly accurate, training on this dataset will likely improve the model performance. 

However, SSFL in the decentralized setting is much more challenging than standard SSL in the centralized setting due to the separate locations of the labeled dataset and the unlabeled datasets. In standard SSL, supervised learning and unsupervised learning are conducted as a single process. Specifically, a single loss function, denoted by $\text{Loss}(w)$ comprising a loss on the labeled data, denoted by $\text{Loss}_L(w)$, and a loss on the unlabeled data, denoted by $\text{Loss}_U(w)$, is used to train the model, i.e.,
\begin{align}
    \text{Loss}(w) = \text{Loss}_L(w) + \alpha\text{Loss}_U(w)
\end{align}
where $\alpha$ is weight parameter. Because the model in the early rounds of training has a low accuracy, the filtered dataset may contain many instances with wrong pseudo-labels. To mitigate the negative effect of wrong pseudo-labels, a small $\alpha$ is used at the beginning so that the model training depends more on the reliable labeled dataset. As the model becomes more accurate over time, a larger $\alpha$ can then be used to take the filtered unlabeled dataset into more consideration. However, SSFL is a different situation because the client only has the loss function on the unlabeled data to perform training. Hence, there is no such a parameter $\alpha$ that can be used to control the emphasis on labeled data or unlabeled data. As a result, the relatively large number of incorrect pseudo-labels produced by the suboptimal global model in the early rounds can significantly damage local training and degrade the obtained local model performance as well as the global model performance after aggregation. In fact, the errors will accumulate and propagate across learning rounds, eventually leading to the failure of learning. 

The root cause of the above problem is that the global model in the early rounds can be overconfident about its prediction result, even though it has not reached convergence. This ``stunbornness'' will be further reinforced if the wrong pseudo-labels are used to train the local models, thereby causing even worse prediction performance in the future rounds. To address this issue, we propose to generate the pseudo-labels using not just a single global model in the current round but many historical global models from the previous rounds. In this way, we hope that the ``stunborness'' of a single model can be softened by the collective opinion of multiple models. We call this method \textit{self-ensemble learning} because the multiple models are all obtained in the same learning process, albeit from different rounds. The idea of using historical model ensembles was also used in \cite{chen2020semi}, whose purpose is to derive smoother predictive distributions and reduce the risk of overconfidence class assignment on out-of-distribution data by individual models. 

Specifically, client $k$ computes the average confidence score vector for each of its instance $i \in \mathcal{D}_k$ using the historical models $\{w^0, w^1, ..., w^t\}$ as follows,
\begin{align}
    \bar{p}^t(x_i) = \frac{1}{t}\sum_{j=1}^t p(x_i; w^j)
\end{align}
Note that client $k$ does not need to actually store all historical models because $\bar{p}^t(x_i)$ can be updated incrementally as
\begin{align}
    \bar{p}^t(x_i) = \frac{t-1}{t}\bar{p}^{t-1}(x_i) + \frac{1}{t}p(x_i; w^t)\label{historical_averaging}
\end{align}
Then, the pseudo-label $y^+_i$ of $x_i$ is generated according to
\begin{align}
    y^+_i = \arg\max_m \bar{p}^t_m(x_i)
\end{align}
whose associated confidence score is $\bar{p}^t_{y^+_i}(x_i)$. Using the confidence threshold vector $\tau^t$ calculated by the server, the client filters its local dataset to construst a filtered dataset:
\begin{align}
    \mathcal{D}^{t,+}_k = \{(x_i, y^+_i); i \in \mathcal{D}_k~ \text{and}~\bar{p}^t_{y^+_i}(x_i) \geq \tau^t_{y^+_i}\}
\end{align}
The condition $\bar{p}^t_{y^+_i}(x_i) \geq \tau^t_{y^+_i}$ ensures that only those instances with sufficiently high confidence are filtered into the dataset. 

\subsubsection{Complementary Negative Learning.}
Self-ensemble learning improves the data filtering performance so that only instances with correct pseudo-labels are more likely to be included into the filtered dataset $\mathcal{D}^{t,+}_k$. However, because the self-ensemble in the initial rounds has only a few historical models, the accuracy of the initial models is low, resulting in a relatively large portion of instances with wrong pseudo-labels being filtered into $\mathcal{D}^{t,+}_k$. To provide more useful information to the supervised learning at the clients, especially in the initial rounds, we employ the idea of \textit{complementary negative learning} \cite{kim2019nlnl}, which was originally proposed for learning problems with noisy labels. The rationale is that although it is hard to classify an instance into the correct class, it is usually much easier to exclude the instance from obviously wrong classes. Thus, by assigning complementary labels, namely classes other than the true class, to the instances, more information can be used to update the local model. Importantly, this information can counter the negative effect of including instances with wrong pseudo-labels in $\mathcal{D}^{t,+}_k$, thereby improving the local training performance. 

To this end, in addition to constructing $\mathcal{D}^{t,+}_k$, we construct a second filtered dataset $\mathcal{D}^{t,-}_k$, which contains data instances with complementary labels. The superscript ``$+$'' indicates that the (pseudo-)labels are supposed to be the true class, while ``$-$'' indicates that the (complementary) labels are supposed to be any class other than the true class. Specifically, for each instance $i\in \mathcal{D}_k$, client $k$ finds the set of classes whose corresponding confidence scores obtained by historical averaging (i.e. eqn. \eqref{historical_averaging}) are lower than a pre-determined small threshold $\theta$, and randomly picks one from this set (unless it is an empty set) to assign it as the complementary label of instance $i$, i.e.
\begin{align}
    y^-_i \in \{m: \bar{p}^t_m(x_i) \leq \theta\}
\end{align}
Thus, the complementary dataset is 
\begin{align}
    \mathcal{D}^{t,-}_k = \{(x_i, y^-_i); i \in \mathcal{D}_k~&\text{and}~\exists m,~\text{s.t.}~\bar{p}^t_m(x_i) \leq \theta~\nonumber\\
    &\text{and}~~\bar{p}^t_{y^+_i}(x_i) < \tau^t_{y^+_i}\}
\end{align}
The condition $\bar{p}^t_{y^+_i}(x_i) < \tau^t_{y^+_i}$ ensures that an instance already in $\mathcal{D}^{t,+}_k$ is not included in $\mathcal{D}^{t,-}_k$.

\subsubsection{Local Model Update.} To update the local model, we consider a client loss function consisting of a positive loss component and a negative loss component,
\begin{align}
    L_k(w) = \lambda L^+_k(w) + L^-_k(w)
\end{align}
where 
\begin{align}
    L^+_k(w) &= \frac{1}{|\mathcal{D}^{t,+}_k|}\sum_{i\in \mathcal{D}^{t,+}_k} l(y^+_i, f(A(x_i); w))\\
    L^-_k(w) &= \frac{1}{|\mathcal{D}^{t,-}_k|}\sum_{i\in\mathcal{D}^{t,-}_k} l(y^-_i, 1-f(x_i; w))
\end{align}
and $\lambda > 0$ is weight parameter. Note that the positive loss component $L^+_k(w)$ is calculated based on the pseudo-labels and the strong data augmentation $A(\cdot)$ on the input data, thereby combining consistency regularization with pseudo-labeling. In FedSEAL, we use RandAugment \cite{cubuk2020randaugment} as the strong data augmentation method. Because in the early rounds, the complementary labeling has a much higher correct rate than the pseudo labeling, the weight $\lambda$ is chosen to be small to reduce the risk due to incorrect pseudo labeling. In later rounds when the historical ensemble generates the pseudo labels at a higher correct rate, $\lambda$ gradually becomes larger so that the client loss function puts more emphasis on the positive loss component. 

With the above defined client loss function, client $k$ updates its local model by executing $E_c$ epochs of mini-batch gradient descent with mini-batch size $B_c$. Each mini-batch update step $b$ performs
\begin{align}
    w^{t,b}_k = w^{t,b-1}_k - \nu_c \nabla_w L_k(w)|w^{t,b}_k
\end{align}
where the initial model $w^{t,0}_k$ is the received global model $w^t$, $\nu_c$ is the client learning rate and $\nabla_w L_k(w)|w^{t,b}_k$ is the gradient of $L_k(w)$ evaluated at $w^{t,b}_k$. The obtained local model $w^t_k$ is then uploaded to the server, completing one round of SSFL. The complete algorithm of FedSEAL is summarized in Algorithm 1. 

\begin{algorithm}[tb]
    \caption{FedSEAL}
    \begin{algorithmic}
    \STATE Server trains the initial model $w^0$ based on its labeled dataset
    \FOR{each FL round $t$}
        \STATE \underline{\textit{Step 1: Server Training}}
        \STATE Model averaging to obtain $\bar{w}^t$
        \STATE Update global model $w^t$
        \STATE Calculate confidence threshold vector $\tau^t$
        \STATE \underline{\textit{Step 2: Model Download}}
        \STATE All clients download global model $w^t$ and confidence threshold vector $\tau^t$ from the server
        \STATE \underline{\textit{Step 3: Client Training}}
        \STATE Sample a subset of clients $\mathcal{K}^t \subseteq \mathcal{N}$
        \FOR{each sampled client $k$}
        \STATE Construct positive filtered dataset $\mathcal{D}^{t,+}_k$
        \STATE Construct negative filtered dataset $\mathcal{D}^{t,-}_k$
        \STATE Update local model $w^t_k$ based on $\mathcal{D}^{t,+}_k$ and $\mathcal{D}^{t,-}_k$
        \ENDFOR
        \STATE \underline{\textit{Step 4: Model Upload}}
        \STATE Sampled clients upload local models $w^t_k, \forall k \in \mathcal{K}^t$ to the server
    \ENDFOR
    \end{algorithmic}
\end{algorithm}

\section{Experiments}
\subsection{Experiment Setup}
We conduct experiments for the image classification task on Fashion-MNIST \cite{xiao2017fashion} and CIFAR10 \cite{krizhevsky2009learning} for both IID and non-IID data distributions. \textbf{Fashion-MNIST}:
The server has a dataset of 500 labeled instances for training. There are 10 clients and each client has 1200 unlabeled instances for training. In each FL round, all clients participate. We use ResNet-18 as the backbone model. 
\textbf{CIFAR10}: We adopt the exact same setting as that in the paper of FedMatch. The server training dataset has 5000 labeled instances. There are 100 clients and each client has 490 unlabeled instances. In each FL round, 5 clients are selected to participate in FL. We use ResNet-9 as the backbone model. More details of the experiment setup and additional experiments on the \textbf{SVHN} dataset \cite{netzer2011reading} can be found in the appendix. 

\textbf{Baselines}. We consider the following baselines in our experiments. \textbf{1) Server-SL}: SL trains a model using only its labeled dataset, disregarding all clients' unlabeled data. We consider this as the \textit{lower bound} for any effective SSL algorithm. \textbf{2) FedAvg-SL}: Assuming all clients' data is labeled, this method trains a model using standard FedAvg. Note that this is not a realizable model in practice and we consider it as the \textit{upper bound} for any SSFL algorithm. \textbf{3) FedAvg-FixMatch}:  Naive combination of FedAvg and FixMatch. \textbf{4) FedAvg-UDA}: Naive combination of FedAvg and UDA. \textbf{5) FedMatch}: a state-of-the-art SSFL algorithm proposed in \cite{jeong2020federated}. \textbf{6) FedRGD}: a state-of-the-art SSFL algorithm proposed in \cite{zhang2020improving}. 

\subsection{Experiment Results}

\begin{table}[]
\caption{Test accuracy on Fashion-MNIST}
\label{tab:fmnist}
\vskip 0.15in
\begin{center}
\begin{small}
\begin{sc}
\begin{tabular}{lcccr}
\toprule
Methods & Non-IID (Dir = 0.1) & IID \\
\midrule
Server-SL &  $78.67\%$ & $80.25\%$  \\
FedAvg-SL & $85.40\%$ & $89.97\%$ \\
FedAvg-FixMatch & $77.84\%$ & $77.35\%$ \\
FedAvg-UDA & $75.02\%$ & $79.95\%$ \\
\textbf{FedSEAL} & $\mathbf{82.63\%}$ & $\mathbf{84.28\%}$ \\
\bottomrule
\end{tabular}
\end{sc}
\end{small}
\end{center}
\vskip -0.1in
\end{table}

\subsubsection{Comparison with naive combinations of FL and SSL methods.} We first compare FedSEAL with methods that naively combine existing FL and SSL methods, namely FedAvg-FixMatch and FedAvg-UDA. This set of experiments are conducted on Fashion-MNIST. For FedSEAL, the initial values of the server learning rate and the client learning rate are set as 0.001 with a decay rate 0.995 and local momentum 0.9, and the complementary threshold $\theta$ for negative learning is 0.05. For FixMatch and UDA, the fixed confidence score threshold is 0.9. Table \ref{tab:fmnist} reports the test accuracy of various methods for both the IID and non-IID cases. As expected, FedAvg-SL achieves the highest accuracy because the client instances are assumed to be labeled, thereby setting the (unrealizable) upper-bound performance. FedSEAL outperforms the realizable lower-bound baseline Server-SL as well as the naive combination methods by a large margin of 4-5\%, demonstrating its ability of effectively utilizing clients' unlabeled data. We note that the test accuracy of FedAvg-FixMatch and FedAvg-UDA is only comparable to or even worse than that of Server-SL, suggesting that a naive combination of SSL and FL does not deliver a satisfactory performance in SSFL problems. This is because many unlabeled data instances can receive incorrect pseudo labels. Using a higher confidence score threshold may improve the pseudo-labeling correct rate but results in much less unlabeled data to be utilized. In the extreme case when the threshold is set to 1, both FedAvg-FixMatch and FedAvg-UDA essentially degenerates to Server-FL. 

\begin{figure*}[tb]
\vskip 0.1in
    \centering
    \begin{minipage}[t]{0.33\linewidth}
        \includegraphics[width=0.95\linewidth]{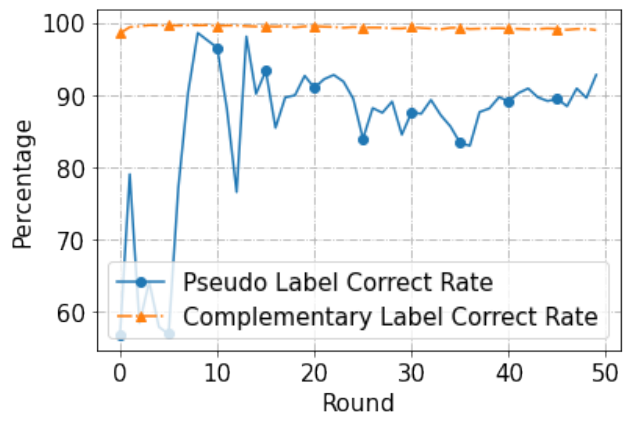}
        \caption{Pseudo-labeling and complementary labeling correct rates}
	    \label{fig:NL_2}
    \end{minipage}
    \begin{minipage}[t]{0.33\linewidth}
        \includegraphics[width=0.95\linewidth]{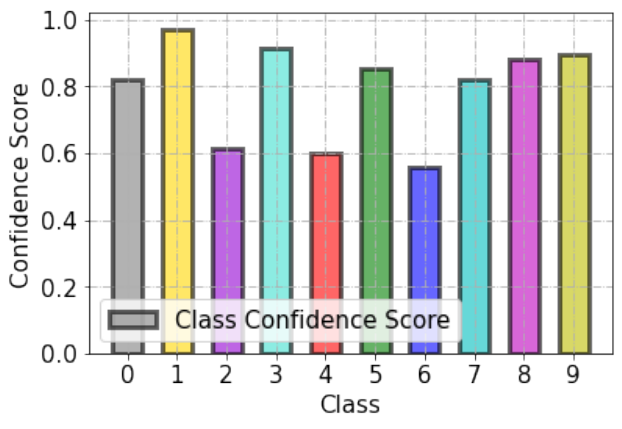}
        \caption{Confidence scores of each class on Fashion-MNIST}
	    \label{fig:con_score}
    \end{minipage} 
    \begin{minipage}[t]{0.33\linewidth}
        \includegraphics[width=0.95\linewidth]{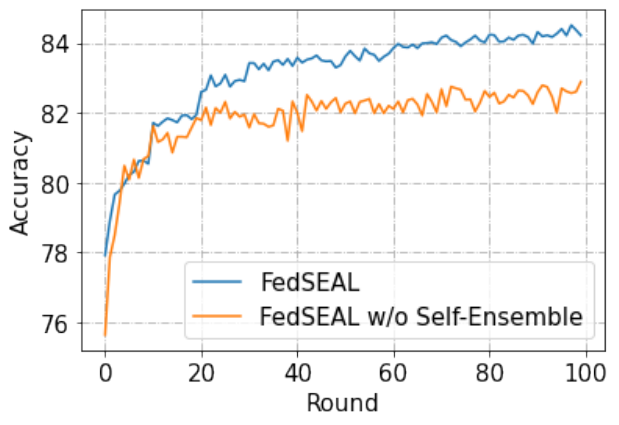}
        \caption{Convergence of FedSEAL without self-ensemble learning}
	    \label{fig:self_ensemble}
    \end{minipage}    
\vskip -0.1in
\end{figure*}

\begin{table}[]
\caption{Test accuracy on CIFAR-10}
\label{tab:cifar}
\vskip 0.15in
\begin{center}
\begin{small}
\begin{sc}
\begin{tabular}{lcccr}
\toprule
Methods & Non-IID & IID \\
\midrule
Server-SL w/o DA & $63.73\%$ & $62.13\%$ \\
Server-SL & $76.90\%$ & $74.30\%$ \\
FedAvg-SL & $86.23\%$ & $88.07\%$ \\
FedMatch & $44.17\%$ & $44.95\%$ \\
FedMatch (corrected) & $56.30\%$ & $54.40\%$ \\
FedRGD & $63.24\%$ & $63.32\%$ \\
\textbf{FedSEAL} & $\mathbf{80.80\%}$ & $\mathbf{81.16\%}$ \\
\bottomrule
\end{tabular}
\end{sc}
\end{small}
\end{center}
\vskip -0.1in
\end{table}

\subsubsection{Comparison with SSFL methods.}
We further compare FedSEAL with state-of-the-art SSFL methods, namely FedMatch and FedRGD. This set of experiments are conducted on CIFAR10, and the setup is exactly the same as that used in FedMatch \cite{jeong2020federated}. Under this same setup, the numbers reported in \cite{jeong2020federated} can be used for a direct comparison. However, during the actual execution of FedMatch by ourselves, we found that these numbers are incorrect due to a wrong parameter configuration, and the actual performance of FedMatch is better than what was reported\footnote{The paper mistakenly stated that 5000 labeled instances were used for the server, but in fact only 1000 labeled instances were used in the experiment, resulting in a lower accuracy. The misreported low accuracy was also used in the comparison for several follow-up works.}. We label the corrected numbers as \textbf{FedMatch (corrected)}. Since FedMatch does not use data augmentation at the server, we further consider an additional baseline that removes data augmentation from Server-SL, labeled as \textbf{Server-SL w/o DA}, for a fair comparison. The parameters of FedSEAL are set the same as the previous experiment. 

Table \ref{tab:cifar} reports the results. Since the lower-bound baseline Server-SL was not used in the comparison for the label-at-server setting in the papers of FedMatch and FedRGD, the results are a little surprising: both FedMatch FedRGD underperform their respective lower-bound baselines (a fair comparison would be FedMatch v.s. Server-SL w/o DA, and FedRGD v.s. Server-SL). By contrast, our FedSEAL algorithm outperforms Server-SL by a large margin, re-confirming its ability of effectively utilizing clients' unlabeled data to boost the learning performance. 

\subsubsection{Impact of negative learning.} In the next sets of experiments, we study the impact of the key components of FedSEAL. We start with negative learning by showing the data filtering performance in the beginning rounds. Fig. \ref{fig:NL_2} shows the percentage of correct pesudo-labels in the postive filtered dataset $\mathcal{D}^{t,+}_k$, and the percentage of correct complementary labels in the negative filtered dataset $\mathcal{D}^{t,-}_k$. As the figure illustrates, the correct rate of the pseudo-labels is considerably lower than that of the complementary labels. Therefore, if supervised learning only uses the positive filtered dataset, its performance can be significantly degraded. 

\subsubsection{Impact of class-wise dynamic confidence threshold.} In FedSEAL, the confidence threshold $\tau$ for data filtering is updated every round and for each class. However, in existing works, similar thresholds are set as a constant and the same for all classes. We chose to design a class-wise dynamic thresholding method because the average confidence scores of different classes can vary significantly. Fig. \ref{fig:con_score} illustrates the confidence scores obtained on the validation dataset by the final global model. For example, if the confidence threshold is set as 0.9 for all classes, then instances in classes 1 and 3 are more likely to pass the filter while those in classes 2, 4 and 6 can hardly pass the filter. Therefore, the filtered positive dataset $\mathcal{D}^{t,+}_k$ can be very or even extremely imbalanced, thereby degrading the model performance. On the other hand, if the confidence threshold is set as 0.5 for all classes, then although instances of all classes can be filtered into $\mathcal{D}^{t,+}_k$, it may contain many instances with wrong labels in classes 1, 3, 5, 7, 8 and 9. 

\subsubsection{Impact of self-ensemble learning.} To show the effectiveness of self-ensemble learning, we replace the average confidence score vector $\bar{p}^t(x_i)$ with the confidence score vector calculated using only the current model $p(x_i;w^t)$ for performing data filtering. Fig. \ref{fig:self_ensemble} illustrates the test accuracy convergence curves for these two methods on Fashion-MNIST. In less than 20 rounds, FedSEAL outperforms FedSEAL w/o self-ensemble learning. This is because as the ensemble size increases, self-ensemble learning helps to filter more unlabeled data with higher quality pseudo-labels into the filtered dataset $\mathcal{D}^{t,+}_k$. 

\section{Conclusion}
FL is a promising ML framework to address the challenges of decentralized datasets and data privacy. However, research on SSFL has not been on par with its SL counterpart. This paper proposed FedSEAL, a new SSFL algorithm, to orchestrate supervised learning at the server and unsupervised learning at the clients. Through extensive experimental validation, FedSEAL is shown to outperform existing SSFL approaches by a large margin, demonstrating the possibility of using unlabeled data to improve learning even in the FL setting. As future work, we plan to investigate how to transfer the techniques adopted by FedSEAL to problems other than image classification, and strive to understand its performance limit from a theoretical perspective.

\bibliographystyle{unsrt}  
\bibliography{references} 

\newpage
\appendix
\onecolumn
\section{Appendix}

\subsection{Experimental Details}


\textbf{Fashion MNIST}: The server has a dataset of 500 labeled instances for training, where each class contains 50 randomly extracted instances (the total number of class is $M = 10$). It also has a validation dataset of 200 instances. There are 10 clients and each client has 1200 unlabeled instances for training. The test dataset has 300 instances for each client. In the IID scenario, each client has 120 randomly extracted unlabeled instances per class whereas in the non-IID scenario, we utilize the Dirichlet function which is typically utilized to simulate the level of non-IID in federated learning. We use ResNet-18 as the backbone model. Other The training hyper-parameters setting of Fashion-MNIST is shown in Table \ref{tab:fmnist_setup}. 

\textbf{CIFAR10}: We adopt the exact same setting as that in the paper of FedMatch \cite{jeong2020federated}. Specifically, a total of 60000 instances in CIFAR10 are split into a training dataset (54000 instances), a validation dataset (3000 instances) and a test dataset (3000 instances). The server training dataset has 5000 labeled instances, where each class contains 500 randomly extracted instances (the total number of class is $M = 10$). There are 100 clients and each client has 490 unlabeled instances. In the IID scenario, each client has 49 randomly extracted unlabeled instances per class whereas in the non-IID scenario, to make a fair comparison, the class distribution is exact same as FedMatch. In each FL round, 5 clients are selected to participate in FL. The only difference is that FedMatch applies a learning rate 0.001 and decays it by a factor of 3 if there is no improvement in the validation loss for 5 consecutive epochs. FedSEAL utilizes the same initial learning rate 0.001 and decays it by 0.995 in each round. The hyper-parameter seting is summarized in Table \ref{tab:cifar_setup}.

\setcounter{table}{0}
\begin{table}[H]
\vskip 0.15in
\begin{center}
\begin{small}
\begin{sc}
\begin{tabular}{lcccr}
\toprule
Hyper-parameters & Values \\
\midrule
    Learning rate & 0.001\\
    Decay rate & 0.995\\
    Momentum & 0.9\\
    Complementary threshold $\theta$ & 0.1\\
    Confidence threshold (Fixmatch) & 0.9\\
    Server batch size & 32\\
    Client batch size & 32\\
    Total number of clients & 10\\
    Number of sampled clients & 10\\
    Number of server local epochs & 5\\
    Number of client local epochs & 5\\
    Positive loss weight $\lambda$ & 0.25\\
    $\lambda$ decay rate (until round = 100) & 0.95\\
    Number of training rounds & 150\\
    \hline
    \end{tabular}
    \caption{Fashion-MNIST setup}
    \label{tab:fmnist_setup}
\end{sc}
\end{small}
\end{center}
\vskip -0.1in
\end{table}

\begin{table}[H]
\vskip 0.1in
    \centering
    \begin{small}
    \begin{sc}
    \begin{tabular}{lcccr}
    
    \toprule
    Hyper-parameters & Values \\
    \midrule
    Learning rate & 0.001\\
    Decay rate & 0.995\\
    Momentum & 0.9\\
    Complementary threshold $\theta$ & 0.05\\
    Server batch size & 100\\
    Client batch size & 100\\
    Total number of clients & 100\\
    Number of sampled clients & 5\\
    Number of server local epochs & 5\\
    Number of client local epochs & 5\\
    Positive loss weight $\lambda$ & 0.1\\
    $\lambda$ decay rate (until round = 100) & 0.95\\
    Number of training rounds & 150\\
    \hline
    \end{tabular}
    \caption{CIFAR10 setup}
    \label{tab:cifar_setup}
\end{sc}
\end{small}
\vskip -0.1in
\end{table}

For the purpose of fair comparison, the backbone model uses ResNet-9, which is used in FedMatch. The architecture of ResNet-9 is given in Table \ref{tab:resnet9}.
\begin{table}[H]
\vskip 0.1in
\begin{small}
    \begin{sc}
    \centering
    \begin{tabular}{lcccr}
\toprule
    Layer & Filter Shape & Stride & Output\\
    \midrule
    Input & N/A & N/A & $32\times32\times3$\\
    Conv 1 & $3\times3\times3\times64$ & 1 & $32\times32\times64$\\
    Conv 2 & $3\times3\times64\times128$ &1 &  $32\times32\times128$\\
    Pool 1 & $2\times2$ & 2 & $16\times16\times128$\\
    Conv 3 & $3\times3\times128\times128$ &1 &  $16\times16\times128$\\
    Conv 4 & $3\times3\times128\times128$ &1 &  $16\times16\times128$\\
    Conv 5 & $3\times3\times128\times256$ &1 &  $16\times16\times256$\\
    Pool 2 & $2\times2$ & 2 & $8\times8\times256$\\
    Conv 6 & $3\times3\times256\times512$ &1 &  $8\times8\times512$\\
    Pool 3 & $2\times2$ & 2 & $4\times4\times512$\\
    Conv 7 & $3\times3\times512\times512$ &1 &  $4\times4\times512$\\
    Conv 8 & $3\times3\times512\times512$ &1 &  $4\times4\times512$\\
    Pool 4 & $4\times4$ & 4 & $1\times1\times512$\\
    \hline
    Softmax & $512\times10$ & N/A & $1\times1\times10$\\
    \hline
    \end{tabular}
    \caption{Network Architecture of ResNet-9}
    \label{tab:resnet9}
\end{sc}
\end{small}
\vskip -0.1in
\end{table}

\subsection{Additional Experiment on Fashion-MNIST}

\subsubsection{A Smaller Backbone Model.} FL is typically run on mobile devices whose GPU resource is limited. Therefore, we also test FedSEAL using a smaller backbone model. In this set of experiments, we replace ResNet-18 with LeNet while keeping the other settings the same. Table \ref{tab:fmnist_lenet} shows that FedSEAL also outperforms the baseline methods across different Non-IID levels.

\begin{table}[H]
\caption{Test accuracy on Fashion-MNIST for LeNet}
\label{tab:fmnist_lenet}
\vskip 0.1in
\begin{center}
\begin{small}
\begin{sc}
\begin{tabular}{lcccr}
\toprule
Methods & Non-IID(0.1) & Non-IID(0.3) & Non-IID(0.9) \\
\midrule
Server-SL &  $74.37\%$ & $74.95\%$ & $74.53\%$  \\
FedAvg-SL & $83.24\%$ & $84.51\%$ & $86.54\%$ \\
FedAvg-FixMatch & $73.34\%$ & $74.73\%$ & $76.00\%$ \\
FedAvg-UDA & $70.68\%$ & $74.23\%$ & $74.91\%$ \\
\textbf{FedSEAL} & $\mathbf{77.90\%}$ & $\mathbf{78.62\%}$ & $\mathbf{79.08\%}$ \\
\textbf{FedSEAL(Strong Augmentation)} & $\mathbf{78.19\%}$ & $\mathbf{78.29\%}$ & $\mathbf{79.14\%}$ \\
\bottomrule
\end{tabular}
\end{sc}
\end{small}
\end{center}
\vskip -0.1in
\end{table}

\subsubsection{Effect of Bootstrapping.}
In FedSEAL, the initial model global model $w^0$ is obtained by supervised learning on the server labeled dataset alone. Fig. \ref{fig:bootstrapping} reflects that compared with applying the random initial global model, utilizing the model which is trained on only the server labeled data can accelerate federated learning convergence and achieve a higher accuracy. This is because as self-ensemble learning in the unsupervised learning part needs to average all historic models to generate pseudo-labels, the earlier-rounds models under random initialization would be less confident in predicting the result and generate unreliable pseudo-labels with a higher probability. 

\subsubsection{The value of complementary threshold $\theta$}
\begin{figure*}[tb]
\vskip 0.1in
    \centering
    \begin{minipage}[t]{0.495\linewidth}
        \includegraphics[width=0.95\linewidth]{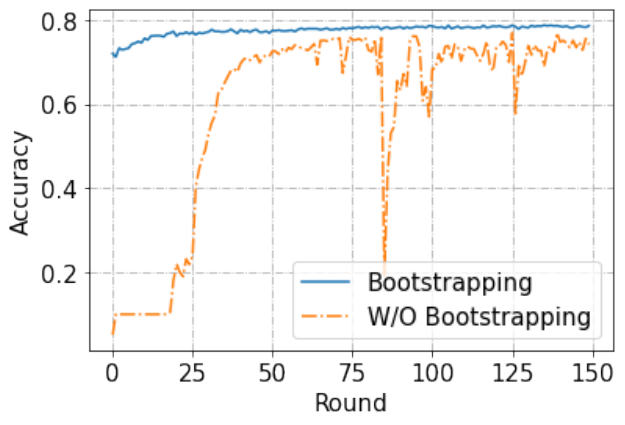}
        \caption{Effect of Bootstrapping on Fashion-MNIST with LENET}
	    \label{fig:bootstrapping}
    \end{minipage}
    \begin{minipage}[t]{0.495\linewidth}
        \includegraphics[width=0.95\linewidth]{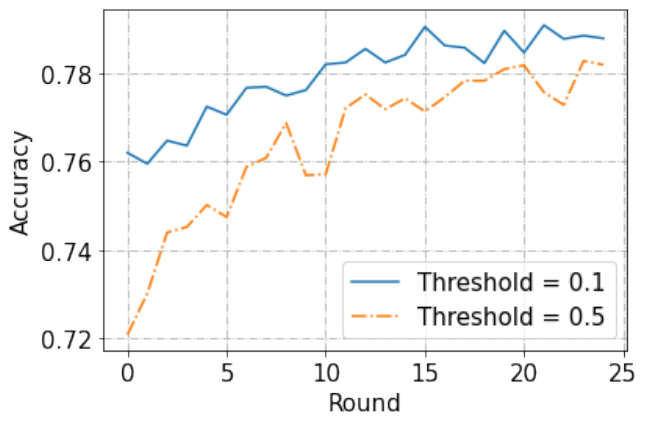}
        \caption{Convergence of first 20 rounds with holding different negative threshold}
	    \label{fig:threshold}
    \end{minipage} 
\vskip -0.1in
\end{figure*}

In the experiments of Fashion-MNIST, we manually set the complementary negative threshold $\theta = 0.1$. Fig. \ref{fig:threshold} illustrates the test accuracy convergence curves in first 25 rounds with different values of complementary threshold. In the first 25 rounds, FedSEAL with $\theta = 0.1$ significantly outperforms the method with $\theta = 0.5$. This is because at beginning rounds, the model predictable ability is still limited which may mistakenly predict a quite low probability for the truth-label class. Hence, exorbitant complementary threshold can enhance the probability of involving unlabeled image with truth-label into complementary negative learning and further decay the model convergence speed.    

\subsubsection{Data augmentation in server supervised learning.}
In the server supervised learning part, we apply random image flipping and random image cropping to perform weak augmentation to enhance the size and quality of server dataset. The result in Table \ref{tab:fmnist_lenet} illustrates utilizing the weak augmentation achieves approximately the same level of test accuracy compared to strong augmentation. To physically reduce the elapsed time, we choose weak augmentation in server supervised learning part.

\subsubsection{Numbers of server labeled data.}
To validate that our method can handle different amounts of labeled data, we conduct the Non-IID (Dir = 0.3) experiments with the number of labeled data instances being 500 or 1000. The results in Table. \ref{tab:fmnist_labeled_data} illustrate that our method FedSEAL can outperform the baseline that trains only on server labeled data, under different numbers of labeled data.

\begin{table}[H]
\caption{Test accuracy on Fashion-MNIST over different numbers of server labeled data}
\label{tab:fmnist_labeled_data}
\vskip 0.1in
\begin{center}
\begin{small}
\begin{sc}
\begin{tabular}{lcccr}
\toprule
Methods & Server with 500 labeled data & Sever with 1000 labeled data \\
\midrule
Server-SL &  $74.95\%$ & $80.17\%$  \\
\textbf{FedSEAL}  & $\mathbf{78.62\%}$ & $\mathbf{83.82\%}$ \\
\bottomrule
\end{tabular}
\end{sc}
\end{small}
\end{center}
\vskip -0.1in
\end{table}

\subsection{Additional Experiment on SVNH.}
To further validate of our method, we conduct additional experiments on the SVHN dataset \cite{netzer2011reading}. We randomly select 50000 instances as the training data, 1000 of which form the labeled dataset at the server and the remaining 49000 are evenly distributed among 100 clients as the unlabeled training dataset. The training setup is described in Table \ref{tab:svhn_setup}.

\begin{table}[]
\vskip 0.1in
    \centering
    \begin{small}
    \begin{sc}
    \begin{tabular}{lcccr}
\toprule
Hyper-parameters & Values \\
\midrule
    Learning rate & 0.001\\
    Decay rate & 0.995\\
    Momentum & 0.9\\
    Complementary threshold $\theta$ & 0.05\\
    Server batch size & 32\\
    Client batch size & 32\\
    Number of total clients & 100\\
    Number of sampled clients & 10\\
    Number of server local epochs & 5\\
    Number of client local epochs & 5\\
    Positive loss weight $\lambda$ & 0.1\\
    $\lambda$ decay rate (until round = 100) & 0.95\\
    Number of training rounds & 150\\
    \hline
    \end{tabular}
    \caption{SVHN setup}
    \label{tab:svhn_setup}
\end{sc}
\end{small}
\vskip -0.1in
\end{table}

The experimental results on SVNH over 3 runs are listed in Table \ref{tab:svnh}.

\begin{table}[H]
\vskip 0.1in
    \centering
    \begin{small}
    \begin{sc}
    \begin{tabular}{lcccr}
    \toprule
    Methods & Non-IID (Dir = 0.1) & IID\\
    \midrule
    Server-SL &  $74.39\pm0.51\%$ & $77.80\pm0.73\%$  \\
    FedAvg-SL & $83.36\pm0.42\%$ & $90.60\pm0.41\%$ \\
    FedAvg-FixMatch & $76.09\pm0.70\%$ & $82.33\pm0.33\%$ \\
    FedAvg-UDA & $75.02\pm0.56\%$ & $80.7\pm0.33\%$ \\
    \textbf{FedSEAL} & $\mathbf{80.00\pm0.58\%}$ & $\mathbf{85.90\pm0.50\%}$ \\
    \hline
    \end{tabular}
    \caption{Test accuracy on SVNH}
    \label{tab:svnh}
\end{sc}
\end{small}
\vskip -0.1in
\end{table}

\subsubsection{Number of Sampled Clients.} We 
change the number of sampled clients and show the performance of FedSEAL on SVNH in Table \ref{tab:svnh_sampled}. The results show that that the number of samples clients does not have a significant impact on the performance.

\begin{table}[H]
\vskip 0.1in
    \centering
    \begin{small}
    \begin{sc}
    \begin{tabular}{lcccr}
    \toprule
    Methods & Class IID\\
    \midrule
    FedMatch (5 sampled clients)& $67.5\pm0.68\%$\\
    FedSEAL (5 sampled clients) & $86.83\pm0.68\%$\\
    FedSEAL (10 sampled clients) & $85.90\pm0.50\%$ \\
    \hline
    \end{tabular}
    \caption{Test accuracy on SVNH over different numbers of sampled clients}
    \label{tab:svnh_sampled}
\end{sc}
\end{small}
\vskip -0.1in
\end{table}

\subsubsection{Hard Labeling in Positive Loss.} In FedSEAL, we combined consistency regularization with pseudo labeling. In this set of experiments, we test the performance of FedSEAL where we replace hard labeling (i.e., $\arg\max_m \bar{p}^t_m(x_i)$) with soft labeling (i.e., $\bar{p}^t_m(x_i)$). Table \ref{tab:svnh_pseudo} shows that the performance degrades by using a soft pseudo-label. The reason for this performance degradation is as follows. Sharpening predictions has been proven beneficial in previous works \cite{xie2019unsupervised,sohn2020fixmatch}. Soft labeling would further soften the class probability distributions in addition to self-ensemble learning. With hard labels, pseudo-labeling is better aligned with cross-entropy minimization, thereby improving the model traning performance.

\begin{table}[H]
\vskip 0.1in
 \begin{small}
    \begin{sc}
    \centering
    \begin{tabular}{lcccr}
    \toprule
    Methods & Class IID\\
    \midrule
    FedSEAL w/o pseudo-labeling (5 sampled clients)& $71.8\pm1.55\%$\\
    FedSEAL (5 sampled clients) & $86.83\pm0.68\%$\\
    \hline
    \end{tabular}
    \caption{Effect of pseudo label in positive loss}
    \label{tab:svnh_pseudo}
\end{sc}
\end{small}
\vskip -0.1in
\end{table}

\subsubsection{A Smaller Backbone Model on SVNH.} Similar to the previous part in the additional experiment on Fashion-MNIST, for SVNH, we replace ResNet-9 with LeNet (we only changed the input dimension to 3) while keeping the other settings the same. Table \ref{tab:svnh_lenet} shows that FedSEAL also significantly outperforms the baseline methods.

\begin{table}[H]
\vskip 0.1in
 \begin{small}
    \begin{sc}
    \centering
    
    \begin{tabular}{lcccr}
    \toprule
    Methods & Non-IID & IID\\
    \midrule
    Server-SL &  $70.43\pm1.65\%$ & $71.17\pm0.31\%$  \\
    FedAvg-SL & $80.33\pm0.54\%$ & $83.50\pm0.73\%$ \\
    FedAvg-FixMatch & $68.36\pm0.78\%$ & $72.57\pm0.66\%$ \\
    FedAvg-UDA & $67.47\pm0.50\%$ & $72.17\pm0.25\%$ \\
    \textbf{FedSEAL} & $\mathbf{75.80\pm0.83\%}$ & $\mathbf{78.93\pm0.39\%}$ \\
    \hline
    \end{tabular}
    \caption{Test accuracy on SVNH for LeNet}
    \label{tab:svnh_lenet}
\end{sc}
\end{small}
\end{table}
\vskip -0.1in

\end{document}